\documentclass[letterpaper, 10 pt, conference]{ieeeconf}
\IEEEoverridecommandlockouts
\usepackage{graphics}           
\usepackage{times}              
\usepackage{amsmath}            
\usepackage{amssymb}            
\usepackage{graphicx}
\usepackage{algorithm}
\usepackage[noend]{algpseudocode}
\usepackage{booktabs}
\usepackage{color}
\usepackage{listings}
\usepackage{subfiles}
\usepackage[acronyms, shortcuts]{glossaries}
\usepackage{hyperref}
\usepackage{subcaption}
\usepackage{makecell}
\usepackage[nocompress]{cite} 
\definecolor{instructioncolor}{rgb}{.5,.5,.5}

\usepackage[font=small]{caption}

\def\secref#1{Sec.~\ref{#1}}
\def\figref#1{Fig.~\ref{#1}}
\def\tabref#1{Tab.~\ref{#1}}
\def\eqref#1{Eq.~(\ref{#1})}

\makeatletter
\usepackage{xspace}
\DeclareRobustCommand\onedot{\futurelet\@let@token\@onedot}
\def\@onedot{\ifx\@let@token.\else.\null\fi\xspace}

\def\etal{{et al}\onedot}
\makeatother

\def\etalcite#1{\etal~\cite{#1}}

\usepackage{array}
\newcolumntype{L}[1]{>{\raggedright\let\newline\\\arraybackslash\hspace{0pt}}m{#1}}
\newcolumntype{C}[1]{>{\centering\let\newline\\\arraybackslash\hspace{0pt}}m{#1}}
\newcolumntype{R}[1]{>{\raggedleft\let\newline\\\arraybackslash\hspace{0pt}}m{#1}}

\def\argmin{\mathop{\rm argmin}}

\newcommand{\RR}{\mathbb{R}}

\renewcommand{\b}[1]{\mbox{\boldmath$#1$}}

\renewcommand{\v}[1]{{\b #1}} 

\newcommand{\m}[1]{{\mbox{{\sffamily\slshape{#1\/}}}}}

\newcommand{\mq}[1]{{\mbox{{\sffamily{#1}}}}}

\newcommand{\cft}[3]{{}^{#1}{#2}_{#3}}

\newcommand{\dvectort}[3]
  {    \left[
            {#1} \; {#2}\; {#3}
       \right] }

\newcommand{\zmatrix}[4]
  {    \left[
          \begin{array}{cc}
            {#1} & {#2} \\
            {#3} & {#4}
          \end{array}
       \right] }

\newcommand{\bb}{\b b}

\newcommand{\bx}{\b x}

\def\lidar{LiDAR}
\def\posegraph{pose-graph}

\newcommand{\SO}[1]{$\mathbb{SO}(#1)$}
\newcommand{\SE}[1]{$\mathbb{SE}(#1)$}
\DeclareRobustCommand{\rchi}{{\mathpalette\irchi\relax}}
\newcommand{\irchi}[2]{\raisebox{\depth}{$#1\chi$}} 

\newacronym{icp}{ICP}{iterative closest point}
\newacronym{slam}{SLAM}{simultaneous localization and mapping}
\newacronym{cticp}{CT-ICP}{continuous time ICP}
\newacronym{imu}{IMU}{inertial measurement unit}
\newacronym{ate}{ATE}{absolute trajectory error}
\newacronym{rpe}{RPE}{relative pose error}

\newcommand{\helipr}{HeLiPR\xspace}
\newcommand{\mulran}{MulRan\xspace}
\newcommand{\apollo}{Apollo\xspace}
\newcommand{\pose}[2]{\cft{#1}{\mq{T}}{#2}}
\newcommand{\posehat}[2]{\cft{#1}{\hat{\mq{T}}}{#2}}
\newcommand{\pert}{\delta \boldsymbol\xi}

\newcommand{\pred}{\hat{\mq{T}}}
\newcommand{\set}[1]{\mathcal{#1}}

\newcommand{\cv}{\Delta \mq{T}_t}

\newcommand{\closurehat}{\posehat{j}{i}}
\newcommand{\grid}{\mathcal{V}}

\newcommand{\locmap}{\set{M}}
\newcommand{\submap}{\set{S}}

\usepackage{flushend}
\makeatletter
\def\blfootnote{\gdef\@thefnmark{}\@footnotetext}
\makeatother

\graphicspath{{pics/}}

\title{\LARGE \bf Improving Map Consistency in Graph-Based LiDAR SLAM\\Through Information-Aware Odometry and Retroactive Loop Closure}

\author{Saurabh Gupta \and Niklas Trekel \and Louis Wiesmann \and Cyrill Stachniss}
\begin{document}
\twocolumn[{
  \vspace{-10pt}
	\renewcommand\twocolumn[1][]{#1}
	\maketitle
	\begin{center}
		\fontsize{9}{9}\selectfont
		\def\svgwidth{0.99\linewidth}
\begingroup%
  \makeatletter%
  \providecommand\color[2][]{%
    \errmessage{(Inkscape) Color is used for the text in Inkscape, but the package 'color.sty' is not loaded}%
    \renewcommand\color[2][]{}%
  }%
  \providecommand\transparent[1]{%
    \errmessage{(Inkscape) Transparency is used (non-zero) for the text in Inkscape, but the package 'transparent.sty' is not loaded}%
    \renewcommand\transparent[1]{}%
  }%
  \providecommand\rotatebox[2]{#2}%
  \newcommand*\fsize{\dimexpr\f@size pt\relax}%
  \newcommand*\lineheight[1]{\fontsize{\fsize}{#1\fsize}\selectfont}%
  \ifx\svgwidth\undefined%
    \setlength{\unitlength}{2804.42773053bp}%
    \ifx\svgscale\undefined%
      \relax%
    \else%
      \setlength{\unitlength}{\unitlength * \real{\svgscale}}%
    \fi%
  \else%
    \setlength{\unitlength}{\svgwidth}%
  \fi%
  \global\let\svgwidth\undefined%
  \global\let\svgscale\undefined%
  \makeatother%
  \begin{picture}(1,0.2518388)%
    \lineheight{1}%
    \setlength\tabcolsep{0pt}%
    \put(0,0){\includegraphics[width=\unitlength,page=1]{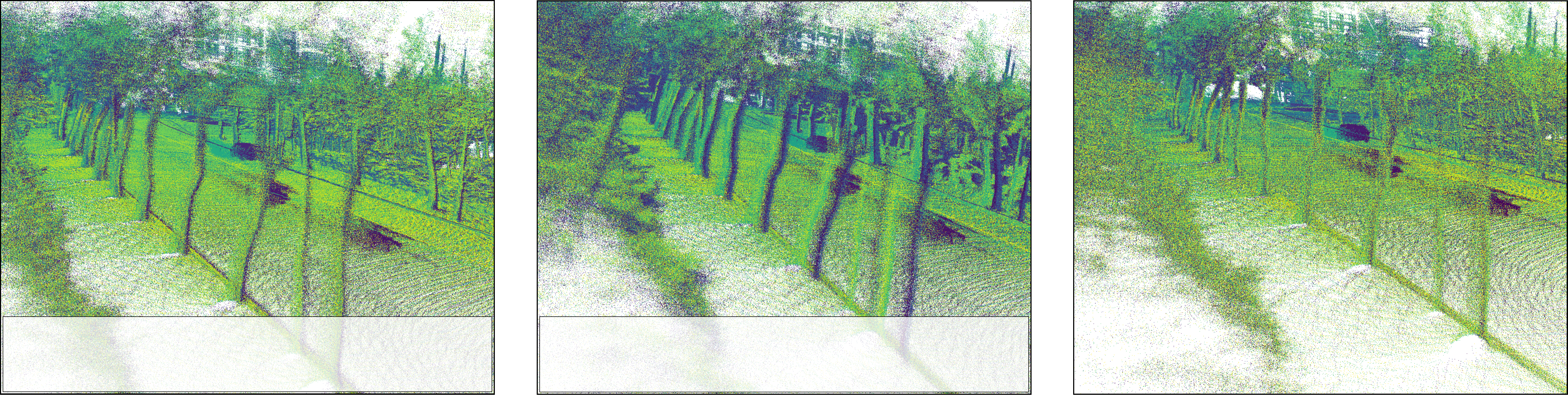}}%
    \put(0.15765705,0.02067951){\color[rgb]{0,0,0}\makebox(0,0)[t]{\lineheight{1.25}\smash{\begin{tabular}[t]{c}KISS-SLAM - ATE: $5.49$\,m\end{tabular}}}}%
    \put(0.49999464,0.02056807){\color[rgb]{0,0,0}\makebox(0,0)[t]{\lineheight{1.25}\smash{\begin{tabular}[t]{c}PIN-SLAM - ATE: $5.25$\,m\end{tabular}}}}%
    \put(0,0){\includegraphics[width=\unitlength,page=2]{motivation.pdf}}%
    \put(0.84246878,0.0205681){\color[rgb]{0,0,0}\makebox(0,0)[t]{\lineheight{1.25}\smash{\begin{tabular}[t]{c}Ours - ATE: $5.51$\,m\end{tabular}}}}%
  \end{picture}%
\endgroup%

  \captionof{figure}{\textbf{Trajectory accuracy does not necessarily imply consistent map quality.} KISS-SLAM\cite{kiss2025iros}, PIN-SLAM~\cite{pan2024tro}, and our approach achieve comparable \ac{ate} values, yet have substantial differences in local geometric consistency. Our method better preserves the alignment of repeated observations, reducing duplicated structures in the reconstructed map. The points in all images are colored based on their timestamp, with the contrast between points reflecting temporally separated revisits.}\label{fig:motivation}
	\end{center}
}]
\thispagestyle{empty}
\pagestyle{empty}

\begin{abstract}
High-quality maps are fundamental for robotics tasks such as navigation and planning. Although modern graph-based LiDAR SLAM systems achieve good trajectory accuracies, a low trajectory error alone does not guarantee geometrically consistent maps, particularly at revisit locations where missed loop closures and residual drift can produce local misalignments.
In this work, we address the problem of jointly improving global trajectory estimation and local map quality in 3D LiDAR SLAM. We first propose a framework to efficiently estimate geometry-dependent information matrices for ICP, enabling principled weighting of odometry constraints in a pose graph. We then introduce a hierarchical loop-closure module that decouples place recognition from geometric registration, together with a retroactive loop-closure module that exploits the optimized pose graph to recover missed loop closures. We also propose an evaluation protocol to measure map consistency at revisit locations.
We evaluate our SLAM system on several datasets against state-of-the-art LiDAR SLAM systems. Experimental results demonstrate global trajectory accuracies on par with or better than existing methods while consistently improving local geometric map consistency at revisit locations. These results suggest that coupling uncertainty-aware odometry with geometry-guided loop-closure refinement leads to more accurate trajectories and higher-quality maps.
\end{abstract}

\blfootnote{All authors are with the University of Bonn, Center for Robotics. Cyrill Stachniss is additionally with the Lamarr Institute for Machine Learning and Artificial Intelligence.%
\indent This work has partially been funded 
by the Deutsche Forschungsgemeinschaft (DFG, German Research Foundation) under Germany's Excellence Strategy, EXC-2070 -- 390732324 -- PhenoRob,
by the Deutsche Forschungsgemeinschaft (DFG, German Research Foundation) under STA~1051/5-1 within the FOR 5351~(AID4Crops),
and by the German Federal Ministry of Research, Technology and Space~(BMFTR) under the Robotics Institute Germany~(RIG).
}%

\section{Introduction}\label{sec:intro}
Mobile robots require accurate maps of their environments for reliable localization and planning. Three-dimensional \lidar{}s have become a popular sensing modality because of their accurate range measurements over large outdoor environments. Consequently, \lidar{}-based \ac*{slam} has become a fundamental component of autonomous robotic systems.

Graph-based \ac{slam} systems~\cite{lu1997ar,grisetti2010titsmag,blanco2025ijrr} combine sensor odometry, loop closure detection, and pose graph optimization to produce globally consistent trajectories. Although these systems achieve good trajectory accuracies, conventional trajectory metrics do not fully capture the geometric consistency of the reconstructed map, see~\figref{fig:motivation}. Small residual pose errors that accumulate over long trajectories, or weakly constrained revisited regions can still result in duplicated surfaces and locally inconsistent geometry despite low global trajectory error.

This limitation arises because pose graph optimization can only exploit the constraints available in the graph. If individual loop closures are missed, revisited regions remain weakly constrained and may still exhibit duplicated surfaces or local geometric inconsistencies even after optimization. Improving map quality therefore requires generating additional geometric constraints in revisited regions.

We address this problem by treating trajectory estimation and map quality as complementary objectives within the \ac{slam} framework. We first improve trajectory estimates through efficient geometry-aware information estimation for \ac{icp} registration, enabling principled weighting of odometry constraints in pose graph optimization. In parallel, we also focus on improving map quality through a hierarchical loop-closure framework followed by a retroactive geometric refinement stage that recovers revisits missed by the appearance-based front-end.

The main contribution of this paper is a novel and effective 3D \lidar{} \ac{slam} approach that improves global trajectory estimates and local map quality by coupling real-time information estimation for point-to-point \ac{icp}, hierarchical loop closure estimation, and geometry-based retroactive loop closure detection within a \posegraph{} optimization framework.

First, we introduce a computationally efficient method for estimating geometry-dependent information matrices for point-to-point \ac{icp} in real time, making information-weighted pose graph optimization practical without modifying the underlying registration algorithm.
Second, we introduce a hierarchical multi-resolution loop-closure framework that decouples place recognition from geometric registration by performing them at different spatial resolutions. Large local maps enhances the robustness of detected loops, while smaller submaps achieve better registration accuracy.
Third, we propose a geometry-based retroactive loop-closure detector that closes the feedback loop between the back-end and the front-end by using the optimized pose graph to recover valid revisits missed during appearance-based place recognition.
Finally, because conventional trajectory metrics do not quantify local geometric consistency, we introduce an evaluation protocol that explicitly measures map quality at revisit locations alongside global trajectory accuracy.

In summary, we make two key claims: Our approach first provides global trajectory accuracy that is on par with or better than state-of-the-art \ac{slam} systems and second, it improves the local map quality in revisited areas, even in scenes with perceptual aliasing. These claims are backed up by the paper and our experimental evaluation.

\section{Related Work}\label{sec:related}
The majority of today's 3D \lidar{} \ac{slam} systems typically follow the graph-based \ac{slam} paradigm~\cite{grisetti2010titsmag,lu1997ar}, consisting of an odometry front-end, loop closure detection, and pose graph optimization. Existing systems differ primarily in their scan registration and map representation, including point-to-point \ac{icp}~\cite{kiss2025iros}, multi-metric registration~\cite{pan2021icra-mvls,zhang2014rss}, continuous-time methods~\cite{dellenbach2022icra}, surfel maps~\cite{behley2018rss}, neural representations~\cite{pan2024tro,deng2023iccv}, or semantic front-ends~\cite{chen2019iros,wang2025iros}. Despite these differences, they all ultimately rely on pose graph optimization to reduce accumulated drift after revisits.

Pose graph optimization assumes appropriately weighted measurement constraints typically expressed through the information matrix of the measurement. While graph optimization supports adaptive and anisotropic information matrices, many systems still rely on fixed isotropic weights as estimating uncertainty for \ac{icp} remains computationally expensive. Although analytical approaches can estimate the \ac{icp} information matrix from the Gauss-Newton Hessian~\cite{brossard2020ral,censi2007icra}, these approximations become inaccurate for point-to-point or other multi-metric \ac{icp} objectives because correspondence changes violate the known-correspondence assumption underlying the Gauss-Newton approximation~\cite{bonnabel2016acc}. Sampling-based approaches estimate uncertainty through repeated perturbation and registration trials~\cite{landry2019icra,nieto2007ras,ma2025ral}, but they are computationally expensive. Real-time information estimation for \ac{icp} therefore remains an obstacle to fully exploiting uncertainty-aware pose graph optimization. We propose a linear regression estimator for the information matrix, fit directly to the samples drawn from the loss landscape, compatible with real-time~\ac{icp} algorithms.

Another important component of a \ac{slam} front-end is a loop-closure detector. Reliable loop closures require both robust place recognition and accurate geometric verification. Practical systems must remain robust under perceptual aliasing while avoiding catastrophic false-positive loop closures~\cite{bosse2004ijrr}. Scan-based loop-closure detectors are used by several \ac{slam} algorithms~\cite{kim2018iros,kim2024ral}. Scan-based methods provide accurate geometric alignment but operate on limited spatial context, making them more susceptible to perceptual aliasing. Practical systems therefore use conservative thresholds, which reduce false positives at the expense of missed loop closures.

An alternative is to perform loop closure using aggregated local maps or submaps instead of individual scans~\cite{gupta2024icra,yuan2023icra,magnusson2009icra-aldf}. Aggregating scans leads to descriptor robustness and naturally produces sparse \posegraph{} constraints. However, larger submaps also accumulate odometry drift, degrading the accuracy of geometric registration. In contrast, scan-level registration is not affected by accumulated drift but provides less spatial context for place recognition. Our framework combines the strengths of both representations through a map hierarchy in the front-end, using large local maps for robust place recognition and smaller submaps for accurate geometric alignment. Existing hierarchical \ac{slam} systems, on the other hand, employ hierarchy primarily in the back-end to improve optimization scalability and efficiency~\cite{grisetti2010icra}.

Conventional graph-based \ac{slam} pipelines treat the front-end and back-end as sequential modules: the front-end generates loop-closure constraints, and the back-end optimizes the pose graph. Since front-end loop closure detection is inherently imperfect, considerable work has focused on making pose graph optimization robust to false-positive constraints through robust kernels~\cite{sunderhauf2012icra}, or max-mixture models~\cite{olson2013ijrr}.

Our work addresses the complementary problem. Rather than increasing the back-end's tolerance to incorrect loop closures, we deliberately employ a conservative front-end to avoid false positives and then use the optimized graph geometry to recover valid loop closures that were initially missed, particularly in perceptually aliased environments. This creates a feedback loop between the front-end and the back-end, and still allows for the use of a robust kernel. Overall, our framework improves both global trajectory accuracy and map quality through information-aware odometry, hierarchical loop-closure front-end, and geometry-guided back-end loop closures.

\begin{figure*}[t]
	\centering
	\def\svgwidth{0.99\linewidth}
	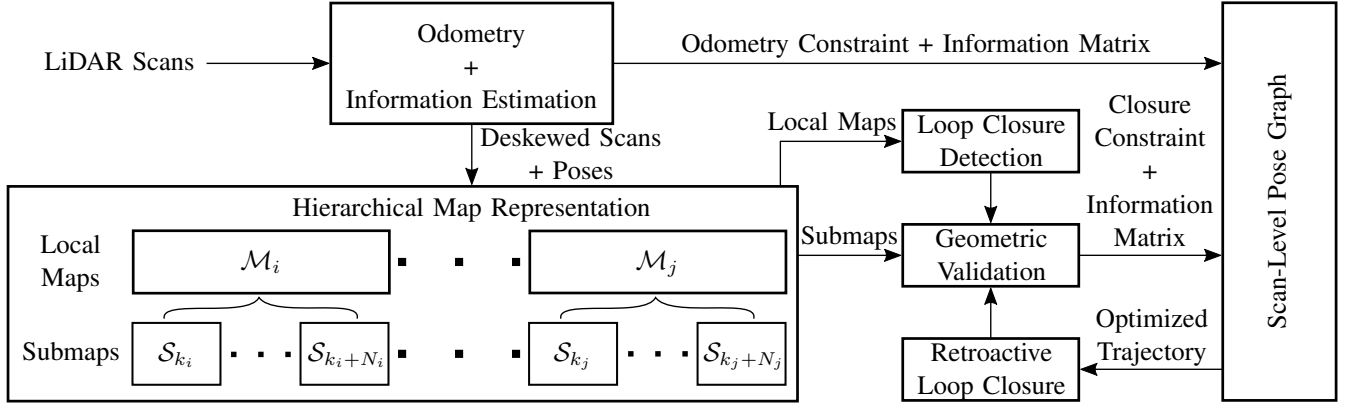
	\caption{\textbf{Overview of our \lidar{} \ac{slam} approach.} The front-end computes \lidar{} odometry and information matrix estimates. Registered scans are organized into a hierarchical map representation with large local maps for place recognition and smaller submaps for geometric validation. Odometry and loop-closure constraints are both optimized in a scan-level pose graph. The optimized trajectory is fed back to a retroactive loop-closure module, which recovers loop closures missed by the appearance-based detector.}
	\label{fig:overview}
  \vspace{-5pt}
\end{figure*}

\section{Our Approach}\label{sec:main}
In this section, we describe the main components of our \ac{slam} pipeline, as illustrated in~\figref{fig:overview}. Given a sequence of \lidar{} scans, we compute the incremental odometry estimates along with an approximate uncertainty estimate for each incremental frame-to-map alignment in~\secref{sec:main_icp} and~\secref{sec:uncertainty}. Then, we introduce the hierarchical front-end setup for loop closure detection in~\secref{sec:main_hierarchical}. We introduce the \posegraph{} back-end in~\secref{sec:posegraph}. Finally, in~\secref{sec:retroactive_closure} we elaborate on the retroactive loop-closure detector in the back-end.

\subsection{\lidar{} Odometry}\label{sec:main_icp}
We use KISS-ICP~\cite{vizzo2023ral} as our \lidar{} odometry front-end; no contribution is claimed for~\secref{sec:main_icp}. Given a point cloud $\set{P}_t = \{\v{p}_i \mid \v{p}_i \in \RR^3\}$ at time~$t$, we create a deskewed and downsampled~$\set{P}_t^* \subset \set{P}_t$. We initialize the pose estimate as~${}^o\pred_t = \pose{o}{t-1} \cv$, where~$\pose{o}{t-1} \in$ \SE{3} is the previous pose estimate,~$\cv \in$ \SE{3} is the constant-velocity motion prediction and~$o$ refers to the odometry frame.

We register~$\set{P}_t^*$ to a voxel local map \mbox{$\set{Q} = \{\v{q}_j \mid \v{q}_j \in \RR^3\}$} by minimizing the point-to-point \ac{icp} objective
\begin{equation}
  \label{eq:icp_cost}
  \rchi(\pose{}{}) = \sum_{(i,j) \in \set{C}(\pose{}{})} \left\|\pose{}{} \v{p}_i - \v{q}_j\right\|^{2}_{2},
\end{equation}
where~$\set{C}$ are the nearest-neighbor matches between~$\set{P}_t^*$ and~$\set{Q}$ for a given pose~$\pose{}{} \in$ \SE{3}. Starting from~${}^o\pred_t$, we solve
\begin{equation}
  \label{eq:icp_optimization}
	\pert = \argmin_{\pert} \rchi({}^o\pred_{t}\boxplus\pert),
\end{equation}
where~$\pert \in \RR^6$ is a pose increment and~$\boxplus$ applies this increment to the current estimate. Upon convergence, we obtain the pose~$\pose{o}{t}$ and insert the scan into the local map.

\subsection{Local Information Matrix Approximation}\label{sec:uncertainty}
KISS-ICP provides the frame-to-map transformation but no uncertainty information. We propose a computationally efficient method to approximate the local information matrix of the transformation. We focus on modelling the local geometry of the \ac{icp} objective around the converged pose~$\pose{o}{t}$.

Instead of relying on a fixed-correspondence Gauss-Newton approximation or repeated registration trials, we directly estimate a local Hessian approximation around~$\pose{o}{t}$. We model the point-to-point \ac{icp} objective with a second-order Taylor-series expansion
\begin{equation}
  \label{eq:taylor}
    \rchi(\pose{o}{t} \boxplus \pert_m) \approx \rchi(\pose{o}{t}) + \pert_m^{\sf T} \v{g} + \frac{1}{2}\pert_m^{\sf T} \m{H} \pert_m,
\end{equation}
where~$\v{g} = \nabla \rchi(\pose{o}{t})$ and~$\m{H} = \nabla^2\rchi(\pose{o}{t})$ denote the local gradient and Hessian, respectively. We retain the gradient term to account for imperfect convergence. Subtracting the constant term~$\rchi(\pose{o}{t})$ yields
\begin{align}
  b_m &= \rchi(\pose{o}{t} \boxplus \pert_m) - \rchi(\pose{o}{t})\\
  \label{eq:perturbation}
      &\approx \pert_m^{\sf T} \v{g} + \frac{1}{2}\pert_m^{\sf T} \m{H} \pert_m.
\end{align}

We sample the \ac{icp} objective around~$\pose{o}{t}$ using perturbations~$\set{U} = \{\pert_m \mid \pert_m \in \RR^6\}$ on the tangent space of \SE{3}. We empirically choose perturbations in~$[-0.1,0.1]$\,m for translation and~$[-0.01,0.01]$\,rad for rotation, large enough to induce correspondence changes while staying inside the convergence basin. For each sample, we recompute correspondences and evaluate~$\rchi(\pose{o}{t} \boxplus \pert_m)$.

Since the coefficients of~$\v{g}$ and~$\m{H}$ appear linearly in~\eqref{eq:perturbation}, we obtain a linear system of the form~$\m{A}\bx = \bb$. To reduce computational cost, we approximate translational and rotational sensitivities independently near convergence and neglect second-order translation-rotation coupling terms. This yields the block diagonal approximation
\begin{equation}
  \label{eq:hessian_split}
  \m{H} = \zmatrix{\m{H}_{\text{tr}}}{\m{0}}{\m{0}}{\m{H}_{\text{ro}}},
\end{equation}
where~$\m{H}_{\text{tr}} \in \RR^{3\times3}$ and~$\m{H}_{\text{ro}} \in \RR^{3\times3}$ denote the translational and rotational Hessian blocks, respectively. We can therefore solve two independent regression problems
\begin{align}
  \m{A}_{\text{tr}}\bx_{\text{tr}} = \bb_{\text{tr}} \qquad \text{and} \qquad \m{A}_{\text{ro}}\bx_{\text{ro}} = \bb_{\text{ro}},
\end{align}
where~$\m{A} \in \RR^{M\times9}$, $\bx \in \RR^9$, $\bb \in \RR^M$, and~$M$ denotes the number of perturbations.

For a translational perturbation $\pert_{\text{tr},m} = \dvectort{x_m}{y_m}{z_m}$, the corresponding row of the design matrix~$\m{A}_{\text{tr}}$ is
\begin{equation}
    \m{A}_{\text{tr},m} = \bigl[x_m,y_m,z_m,\tfrac{x_m^2}{2},\tfrac{y_m^2}{2},\tfrac{z_m^2}{2},x_m y_m,y_m z_m,x_m z_m\bigr],
\end{equation}
with unknown parameter vector
\begin{equation}
    \bx_{\text{tr}}^{\sf T} = \left[g_x,g_y,g_z,h_{xx},h_{yy},h_{zz},h_{xy},h_{yz},h_{xz}\right].
\end{equation}

Here, $\dvectort{g_x}{g_y}{g_z}$ denote the translational gradient~$\v{g}_{\text{tr}}^{\sf T}$ and~$h_{ij}$ the six coefficients of the symmetric Hessian block~$\m{H}_{\text{tr}}$.

We construct~$\m{H}_{\text{ro}}$ analogously using rotational perturbations on the tangent space of \SO{3}. We then combine~$\m{H}_{\text{tr}}$ and~$\m{H}_{\text{ro}}$ as per~\eqref{eq:hessian_split} to obtain~$\m{H}$. We project~$\m{H}$ onto a positive-definite cone through eigendecomposition, where we clamp all eigenvalues below~$10^{-3}$ to~$10^{-3}$ to obtain $\m{H}^+$.

Following the standard interpretation of curvature as information in nonlinear least-squares estimation, we use~$\Lambda_{\text{odo}} = \m{H}^+$ as the information matrix for the odometry constraint in the pose graph optimization. Note that this information estimation framework can also be used with other \ac{icp} objectives without loss of generality.

\subsection{Hierarchical Front-End}\label{sec:main_hierarchical}
We maintain a hierarchical map representation in the front-end, as illustrated in~\figref{fig:overview}. Large local maps provide sufficient spatial context for reliable place recognition. Each local map consists of multiple smaller submaps that accumulate less odometry drift and are therefore used for geometric registration. We generate both representations online using a displacement-based splitting criterion.

We create a submap whenever the platform displaces more than~$\tau_{\text{sub}}=25$\,m. A submap is represented as $\submap_k = (\pose{m}{k}^{\text{sub}}, \grid_k^{\text{sub}}, \Gamma_k^{\text{sub}})$, where~$\pose{m}{k}^{\text{sub}}$ denotes the keypose corresponding to the first scan,~$\grid_k^{\text{sub}}$ is a voxel grid with resolution~$\nu_{\text{sub}}$ and~$\Gamma_k^{\text{sub}} = (\pose{t_k^s}{t_k^s}, \ldots, \pose{t_k^s}{t_k^e})$ is the local trajectory segment, where $t_k^s$ and~$t_k^e$ denote the timestamps of the first and the last scan of the~$k^{\text{th}}$ submap, respectively.

In parallel, we create local maps using a larger displacement threshold~$\tau_{\text{loc}} = 100$\,m. To maintain an unambiguous association between submaps and local maps, every local-map split simultaneously triggers a submap split. This ensures that each submap belongs to exactly one local map. A local map is represented as \mbox{$\locmap_n = (\pose{m}{n}^{\text{loc}},~\grid_n^{\text{loc}}, \set{I}_n)$}, where~$\pose{m}{n}^{\text{loc}}$ denotes the keypose corresponding to the first scan,~$\grid_n^{\text{loc}}$ is a voxel grid with resolution~$\nu_{\text{loc}}$ and~\mbox{$\set{I}_n = \{k_n, \ldots, k_n + N_n\}$} are the indices of its constituent submaps. We also define the local-map trajectory by concatenating the trajectories of its constituent submaps,
\begin{equation}
  \Gamma_{n}^{\text{loc}} = \text{concat}\left(\pose{m}{k_n}^{\text{sub}}\Gamma_{k_n}^{\text{sub}}, \ldots, \pose{m}{k_n + N_n}^{\text{sub}}\Gamma_{k_n + N_n}^{\text{sub}}\right),
\end{equation}
where transforming a trajectory by a pose denotes left multiplication of every pose in the trajectory.

The place-recognition stage, based on the method of Gupta~\etalcite{gupta2026ijrr}, identifies two overlapping local maps and provides an approximate relative pose. For a query local map~$\locmap_i$, we search for the local map~$\locmap_j$ with the largest number of inlier feature matches exceeding~$\tau_{\text{inliers}}$. We accept such a pair~$(i,j)$ of local maps as a loop closure candidate and get an initial alignment~$\closurehat \in$ \SE{3} between their keyposes~$\pose{m}{i}^{\text{loc}}$ and~$\pose{m}{j}^{\text{loc}}$ from the feature matches.

Since each local map contains multiple submaps within the hierarchical structure, we next identify the pairs of submaps that observe the same physical region. Only those pairs are used to generate loop-closure constraints. We identify such pairs using trajectory proximity instead of voxel map-based overlap because it is computationally cheaper yet provides a reliable estimate of spatial overlap between submaps.

For each source submap with index~$k \in \set{I}_i$ and target submap with index~$r \in \set{I}_j$, we express their keyposes relative to their parent local-map coordinate frame as
\begin{equation}
  \cft{i}{\tilde{\mq{T}}}{k}^{\text{sub}} = (\pose{m}{i}^{\text{loc}})^{-1}\,\pose{m}{k}^{\text{sub}}\quad\text{and}\quad\cft{j}{\tilde{\mq{T}}}{r}^{\text{sub}} = (\pose{m}{j}^{\text{loc}})^{-1}\,\pose{m}{r}^{\text{sub}}.
\end{equation}

Using the initial loop closure estimate~$\closurehat$, we transform the source submaps into the target local-map frame as~$\cft{j}{\hat{\mq{T}}}{k}^{\text{sub}} = \closurehat \cft{i}{\tilde{\mq{T}}}{k}^{\text{sub}}$. We then transform the source and target submap trajectories as
\begin{equation}
  \hat{\Gamma}_k^{\text{sub}} = \cft{j}{\hat{\mq{T}}}{k}^{\text{sub}}\Gamma_k^{\text{sub}}\quad\text{and}\quad\tilde{\Gamma}_r^{\text{sub}} = \cft{j}{\tilde{\mq{T}}}{r}^{\text{sub}}\Gamma_r^{\text{sub}}.
\end{equation}

This expresses both submap trajectories in a common coordinate frame of the target local map~$j$, enabling direct geometric comparison.

We identify corresponding submaps by comparing these transformed trajectories using the Hausdorff distance
\begin{equation}
  \label{eq:hausdorff}
  \begin{split}
    d_{H}\left(\Gamma_1, \Gamma_2\right) = \max\left( \max_{\v{t}_a \in \Gamma_1}\left(\min_{\v{t}_b \in \Gamma_2}d\left(\v{t}_a,\v{t}_b\right)\right),\right.\\
    \left.\max_{\v{t}_b \in \Gamma_2}\left(\min_{\v{t}_a \in \Gamma_1}d\left(\v{t}_a,\v{t}_b\right)\right)\right),
  \end{split}
\end{equation}
where~$d\left(\v{t}_a, \v{t}_b\right)$ is the L2 norm between the translational components~$\v{t}_a$ and~$\v{t}_b$ of the corresponding trajectories.

For each source submap~$k \in \set{I}_i$, we identify the nearest target submap using~\eqref{eq:hausdorff} as
\begin{equation}
  r^*(k) = \argmin_{r \in \set{I}_j} d_{\text{H}} \left(\hat{\Gamma}_k^{\text{sub}}, \tilde{\Gamma}_r^{\text{sub}}\right),
\end{equation}
and accept the pair~$(k,r^*(k))$ whenever
\begin{equation}
  d_{\text{H}}\left(\hat{\Gamma}_k^{\text{sub}}, \tilde{\Gamma}_{r^*(k)}^{\text{sub}}\right) \leq \tau_{\text{H}},
\end{equation}
where~$\tau_{\text{H}}$ denotes the Hausdorff distance threshold, yielding the candidate correspondence set
\begin{equation}
  \set{Z}_{ij} = \left\{(k,r^*(k)) \mid k \in \set{I}_i, d_{\text{H}}\left(\hat{\Gamma}_k^{\text{sub}}, \tilde{\Gamma}_{r^*(k)}^{\text{sub}}\right) \leq \tau_{\text{H}}\right\}.
\end{equation}

Finally, for every candidate pair~$(k, r) \in \set{Z}_{ij}$, we perform fine registration between~$\grid_k^{\text{sub}}$ and~$\grid_r^{\text{sub}}$. We initialize the fine registration with
\begin{equation}
  \posehat{r}{k} = (\cft{j}{\tilde{\mq{T}}}{r}^{\text{sub}})^{-1}\,\cft{j}{\hat{\mq{T}}}{k}^{\text{sub}},
\end{equation}
yielding a refined relative pose~$\pose{r}{k}$, the associated information matrix~$\Lambda_{\text{lc}}$, and a registration fitness score~$s_{rk}$, where larger fitness scores indicate better alignment. If the registration fitness score~$s_{rk}$ exceeds the acceptance threshold~$\tau_{\text{f}}$, we insert the loop-closure constraint into the pose graph.

\subsection{Pose Graph Back-End}\label{sec:posegraph}
We maintain a pose graph~$\set{G} = (\set{V}, \set{E})$ at the back-end of our pipeline, where~$\set{V}$ is the set of nodes and~\mbox{$\set{E} = \set{E}_{\text{odo}} \cup \set{E}_{\text{lc}}$} consists of sequential odometry constraints~$(\set{E}_{\text{odo}})$ and loop-closure constraints~$(\set{E}_{\text{lc}})$. We employ a scan-level pose graph representation to enable smooth deformation of the estimated global trajectory, with errors distributed over each scan in the graph upon the insertion of a loop closure. As we detect loop closure at most once per local map~(every $100$\,m), pose graph optimization is performed infrequently and therefore does not affect the runtime of our SLAM algorithm.

Each incoming \lidar{} scan is a node in the pose graph. For every scan acquired at time~$t$ belonging to submap~$k$, we obtain its pose in the global map frame~$\pose{m}{t} = \pose{m}{k}^{\text{sub}}\mq{T}_{t_k^s\rightarrow t}$, where~$t \in [t_k^s, t_k^e]$. We connect consecutive scan nodes by relative odometry constraints~$(\set{E}_{\text{odo}})$ along with their corresponding information matrices~$\Lambda_{\text{odo}}$. After detecting and validating loop closures between submaps belonging to local maps~$\locmap_i$ and~$\locmap_j$, we add a loop-closure constraint~$(\set{E}_{\text{lc}})$ between the nodes corresponding to the first scan in the submaps with the corresponding information matrix~$\Lambda_{\text{lc}}$.

After a pose graph optimization, we propagate the optimized poses back to the hierarchical map representation. For every submap, we update its keypose with the optimized pose of the node corresponding to its first scan and recompute the complete trajectory segment from the subsequent nodes.

\subsection{Geometry-Based Retroactive Loop Closure Detection}\label{sec:retroactive_closure}
The front-end loop-closure detector used in this work~\cite{gupta2026ijrr} favors precision over recall through a self-similarity feature-pruning stage in order to avoid false positives, particularly in perceptually aliased environments. While this conservative strategy improves robustness and avoids wrong loop closures, it inevitably misses valid loop closures, leaving portions of the global trajectory weakly constrained, potentially producing locally imperfect maps.

However, once the robust loop closures are incorporated, the pose graph optimization redistributes the accumulated drift, improving the global alignment of revisited regions. Trajectories corresponding to the same physical location, previously separated in the estimated map due to accumulated drift are now in closer geometric proximity. Exploiting this observation, we perform a retroactive loop closure based on geometric proximity, recovering loop closures from areas where appearance-based loop closure acts conservatively.

In particular, we revisit previously created local maps and identify candidate loop closures in the local vicinity by comparing their optimized trajectories. We could naïvely revisit every local map after each optimization, but this would be computationally expensive. Instead, we revisit only the contiguous sequence of local maps that lie between the current local map and the previous local map to have established a successful appearance-based loop closure. These maps have not previously benefited from loop-closure corrections, making them the most promising candidates for recovering additional loop closures after optimization.

We compute the Hausdorff distance, see~\eqref{eq:hausdorff}, between the optimized trajectories of each candidate local-map pair. We retain a local-map pair~$(i,j)$ if $d_{\text{H}}\left(\Gamma_i^{\text{loc}}, \Gamma_j^{\text{loc}}\right) \leq \tau_{\text{H}}$. This coarse geometric search substantially reduces the number of submap pairs that require fine registration.

We then pass the retained pairs to the hierarchical submap association module described in~\secref{sec:main_hierarchical}. The optimized pose graph provides an accurate initial estimate of the relative pose between nearby submaps as
\begin{equation}
  \posehat{r}{k} = (\pose{m}{r}^{\text{sub}})^{-1}\,\pose{m}{k}^{\text{sub}},
\end{equation}
which we use directly to initialize the fine registration.

Consequently, the retroactive loop-closure detector exploits the geometric consistency of the optimized pose graph, recovering loop closures that the front-end intentionally rejected or failed to detect. This creates a feedback loop between the front-end and the back-end by using the optimized graph to generate new front-end loop closure associations.

\section{Experimental Evaluation}\label{sec:exp}
The main focus of this work is an effective 3D \lidar{} \ac{slam} approach that improves global trajectory estimation and local map quality through information estimation-aware odometry, appearance-based front-end loop closure, and geometry-based retroactive loop closure.

We evaluate our system on several large-scale benchmark datasets and compare it against state-of-the-art \lidar{} \ac{slam} systems. The experiments also support our key claims, that our system
(i) provides global trajectory accuracy that is on par with or better than state-of-the-art \ac{slam} systems;
(ii) improves the local map quality in revisited areas, even in scenes with perceptual aliasing.

\begin{table*}[t]
    \caption{Quantitative results for global trajectory accuracy. We report the ATE in meters and RPE in percentage as [m]\,/\,[\%]. The best and second-best methods are in \textbf{bold} and \underline{underline}, respectively. We provide ablation studies at the bottom of each table.}
    \centering
    \begin{subtable}[t]{0.99\linewidth}
        \caption{\centering \helipr{} dataset}\label{tab:AvgHeLiPR}
        \centering
        \begin{tabular}{lccccccccc}
            \toprule
            Method & \makecell{Bridge \\ Aeva} & \makecell{Bridge \\ Avia} & \makecell{Bridge \\ Ouster} & \makecell{Roundabout \\ Aeva} & \makecell{Roundabout \\ Avia} & \makecell{Roundabout \\ Ouster} & \makecell{Town \\ Aeva} & \makecell{Town \\ Avia} & \makecell{Town \\ Ouster} \\ 
            \midrule
            MULLS & 356.06\,/\,9.17 & 321.87\,/\,4.06 & 52.65\,/\,0.68 & 19.08\,/\,1.43 & 16.39\,/\,1.06 & 2.65\,/\,0.24 & 39.82\,/\,2.85 & 14.93\,/\,1.35 & 4.45\,/\,0.27 \\ 
            CT-ICP & -\,/\,- & \underline{41.47}\,/\,\underline{0.36} & -\,/\,- & 10.04\,/\,0.72 & \underline{3.36}\,/\,\underline{0.25} & 1.81\,/\,\underline{0.16} & 65.29\,/\,1.37 & 63.72\,/\,2.02 & -\,/\,- \\ 
            PIN-SLAM & -\,/\,- & 365.72\,/\,2.15 & \underline{19.23}\,/\,\textbf{0.13} & -\,/\,- & 7.02\,/\,0.51 & 1.47\,/\,\textbf{0.11} & 41.19\,/\,1.03 & 11.40\,/\,0.64 & 2.55\,/\,\textbf{0.12} \\ 
            KISS-SLAM & \underline{98.61}\,/\,1.57 & 148.88\,/\,1.82 & 19.47\,/\,0.28 & \underline{6.06}\,/\,\underline{0.47} & 3.84\,/\,\textbf{0.23} & \textbf{1.18}\,/\,0.17 & \underline{14.44}\,/\,\underline{0.85} & 12.01\,/\,\underline{0.46} & \underline{1.99}\,/\,0.21 \\ 
            Ours & \textbf{16.83}\,/\,\textbf{0.47} & \textbf{20.17}\,/\,\textbf{0.25} & \textbf{14.18}\,/\,\underline{0.22} & \textbf{4.67}\,/\,\textbf{0.46} & \textbf{3.34}\,/\,\textbf{0.23} & \underline{1.21}\,/\,0.18 & \textbf{10.50}\,/\,\textbf{0.47} & \textbf{7.57}\,/\,\textbf{0.38} & \textbf{1.89}\,/\,\underline{0.20} \\ 
            \midrule
            \multicolumn{10}{c}{Ablation Study: no-info $\rightarrow$ without uncertainty estimation module; no-feedback $\rightarrow$ without the retroactive loop closing} \\
            \midrule
            no-info & 107.34\,/\,1.46 & 27.03\,/\,0.39 & 18.18\,/\,0.29 & 6.01\,/\,0.44 & 3.93\,/\,0.25 & 1.31\,/\,0.20 & 11.78\,/\,0.83 & 8.70\,/\,0.59 & 1.94\,/\,0.22 \\ 
            no-feedback & 16.37\,/\,0.47 & 21.33\,/\,0.25 & 13.80\,/\,0.21 & 4.43\,/\,0.39 & 3.89\,/\,0.22 & 1.14\,/\,0.18 & 13.47\,/\,0.48 & 8.31\,/\,0.38 & 1.75\,/\,0.20 \\ 
            \bottomrule
        \end{tabular}
        \vspace{10pt}
    \end{subtable}
    \begin{subtable}[t]{0.99\linewidth}
        \caption{\centering \apollo{} and \mulran{} dataset}\label{tab:ApolloMulRan}
        \centering
        \begin{tabular}{lcccccccc}
            \toprule
            Method & \makecell{\apollo{} \\ BTS} & \makecell{\apollo{} \\ CP} & \makecell{\apollo{} \\ H237} & \makecell{\apollo{} \\ MAVE} & \makecell{\apollo{} \\ SB} & \makecell{\mulran{} \\ KAIST} & \makecell{\mulran{} \\ Riverside} & \makecell{\mulran{} \\ Sejong} \\ 
            \midrule
            MULLS & 104.14\,/\,0.27 & 47.03\,/\,0.90 & 354.21\,/\,\underline{0.39} & 182.59\,/\,\textbf{0.16} & -\,/\,- & 34.21\,/\,0.60 & 66.60\,/\,2.27 & 1082.55\,/\,2.53 \\ 
            CT-ICP & 10.81\,/\,0.27 & 0.97\,/\,0.30 & 258.87\,/\,0.62 & 61.39\,/\,0.33 & 667.28\,/\,1.60 & 2.76\,/\,0.50 & 8.37\,/\,0.58 & -\,/\,- \\ 
            PIN-SLAM & 6.10\,/\,0.24 & \textbf{0.64}\,/\,\textbf{0.16} & 97.11\,/\,1.66 & \underline{7.06}\,/\,0.20 & \underline{2.23}\,/\,\underline{0.25} & \textbf{2.42}\,/\,\textbf{0.32} & \underline{7.55}\,/\,0.46 & 782.96\,/\,7.68 \\ 
            KISS-SLAM & \textbf{3.74}\,/\,\underline{0.23} & \underline{0.80}\,/\,0.22 & \underline{30.16}\,/\,0.40 & 9.08\,/\,0.26 & 2.51\,/\,\underline{0.25} & 2.96\,/\,\underline{0.33} & 8.34\,/\,\underline{0.41} & \underline{178.35}\,/\,\textbf{0.25} \\ 
            Ours & \underline{4.22}\,/\,\textbf{0.22} & 0.81\,/\,\underline{0.19} & \textbf{26.50}\,/\,\textbf{0.29} & \textbf{6.17}\,/\,\underline{0.19} & \textbf{1.61}\,/\,\textbf{0.17} & \underline{2.67}\,/\,\textbf{0.32} & \textbf{7.22}\,/\,\textbf{0.35} & \textbf{169.14}\,/\,\underline{0.28} \\ 
            \midrule
            \multicolumn{9}{c}{Ablation Study: no-info $\rightarrow$ without uncertainty estimation module; no-feedback $\rightarrow$ without the retroactive loop closing} \\
            \midrule
            no-info & 9.45\,/\,0.23 & 1.02\,/\,0.22 & 88.25\,/\,0.32 & 19.38\,/\,0.19 & 22.13\,/\,0.45 & 2.79\,/\,0.31 & 10.04\,/\,0.63 & 173.44\,/\,0.28 \\ 
            no-feedback & 3.83\,/\,0.22 & -\,/\,- & 27.31\,/\,0.30 & 5.33\,/\,0.20 & 1.94\,/\,0.19 & 2.64\,/\,0.32 & 7.25\,/\,0.35 & 169.16\,/\,0.28 \\ 
            \bottomrule
        \end{tabular}
    \end{subtable}
\end{table*}

\begin{table}[t]
    \centering
    \caption{Quantitative results on the Newer College dataset. We report the ATE in meters and RPE in percentage as [m]\,/\,[\%]. The best and second-best methods are in \textbf{bold} and \underline{underline}, respectively. We provide ablation studies at the bottom of the table.}
    \begin{tabular}{lcc}
        \toprule
        Method     & \makecell{01-short-sequence} & \makecell{02-long-sequence} \\
        \midrule
        MULLS      & 0.47\,/\,\textbf{0.17}             & 8.47\,/\,4633.62                      \\
        CT-ICP     & 0.63\,/\,0.42                      & 25.06\,/\,3.01                        \\
        PIN-SLAM   & 0.42\,/\,0.33                      & \textbf{0.31}\,/\,\textbf{0.22}       \\
        KISS-SLAM  & \textbf{0.30}\,/\,\underline{0.26} & 1.58\,/\,2.14                         \\
        Ours       & \underline{0.34}\,/\,0.38          & \underline{0.42}\,/\,\underline{0.46} \\
        \midrule
        \multicolumn{3}{c}{Ablation Study} \\
        \midrule
        no-info         & 0.36\,/\,0.43                    & 0.42\,/\,0.48                         \\
        no-feedback     & 0.34\,/\,0.38                    & 0.42\,/\,0.46                         \\
        \bottomrule
    \end{tabular}
    \label{tab:NCD}
    \vspace{-5pt}
\end{table}

\subsection{Experimental Setup}
We evaluate our \lidar{} \ac{slam} approach on the \mulran{}~\cite{kim2020icra}, \helipr{}~\cite{jung2024ijrr}, and \apollo{}~\cite{huang2018cvprws} datasets. The \helipr{} dataset is particularly challenging, with long sequences in urban scenarios, driving through highways, narrow alleys, and perceptually aliased environments. It also provides data from multiple \lidar{} sensors, allowing us to evaluate the robustness of \ac{slam} algorithms to different scanning patterns and fields of view. Additionally, we also report performance metrics on the Newer College dataset~\cite{ramezani2020iros}, which is captured with a handheld \lidar{} device.

We compare our approach against several state-of-the-art \lidar{} \ac{slam} systems, including CT-ICP~\cite{dellenbach2022icra}, MULLS~\cite{pan2021icra-mvls}, PIN-SLAM~\cite{pan2024tro}, and KISS-SLAM~\cite{kiss2025iros}. For all baselines, we use the authors' publicly available implementations, with the recommended dataset-specific configurations.

Unless stated otherwise, all experiments use the same parameter set across all datasets:
$\nu_{\text{sub}}=0.5$ m,
$\nu_{\text{loc}}=0.5$ m,
$\tau_{\text{inliers}}=5$,
$\tau_{\text{H}}=100$ m,
and
$\tau_{\text{f}}=0.25$.
We perform no dataset- or sensor-specific parameter tuning.

\subsection{Pose Accuracy Evaluation}
We first evaluate global trajectory accuracy using the Absolute trajectory error~(ATE) and Relative pose error~(RPE) following the KITTI metric~\cite{geiger2013ijrr}. Since the \helipr{} and \mulran{} datasets contain three recordings per environment, we report the average metric over the corresponding sequences.

We present the quantitative results in~\tabref{tab:AvgHeLiPR}, \tabref{tab:ApolloMulRan} and \tabref{tab:NCD}. Overall, our method achieves global trajectory accuracy that is competitive with state-of-the-art baseline approaches across all datasets with the same set of system parameters. On the \helipr{} dataset~\tabref{tab:AvgHeLiPR}, we obtain the lowest \ac{ate} score on the majority of the sequences and also consistently maintain competitive RPE values.

We observe the largest improvements on the challenging Bridge sequences across all three \lidar{}s (Aeva, Avia, and Ouster), which contain long traversals through structurally repetitive environments that are prone to perceptual aliasing. These results suggest that the conservative loop-closure strategy together with geometry-aware constraint weighting increases robustness in such scenarios.

On the \apollo{} and \mulran{} datasets~\tabref{tab:ApolloMulRan}, our approach obtains the best ATE on five of the eight sequences. On the Newer College dataset~\tabref{tab:NCD}, our method's performance matches the strongest baselines despite the significantly different motion profile of the handheld sensing platform. These results indicate that our approach generalizes well across different \lidar{} sensors and operating environments.

The ablation study further illustrates the roles of the individual components. Removing geometry-aware information weighting consistently degrades global trajectory accuracy across most sequences, demonstrating the benefit of information-weighted pose graph optimization. Disabling retroactive loop closure has only a minor effect on ATE, indicating that its primary contribution is not improved localization accuracy but improved local map consistency, which we evaluate in the following experiment. An exception is the narrow field-of-view Avia sensor, where retroactive loop closures yield more accurate trajectories. The reduced field-of-view makes appearance-based place recognition more difficult, allowing the geometry-guided recovery stage to successfully identify additional valid loop closures.

Overall, these quantitative results support our first claim that our \ac{slam} system provides on par with or better than state-of-the-art global trajectory accuracy.
\subsection{Revisit Map Consistency Evaluation}

\begin{figure*}[t]
	\centering
	\def\svgwidth{0.99\linewidth}
	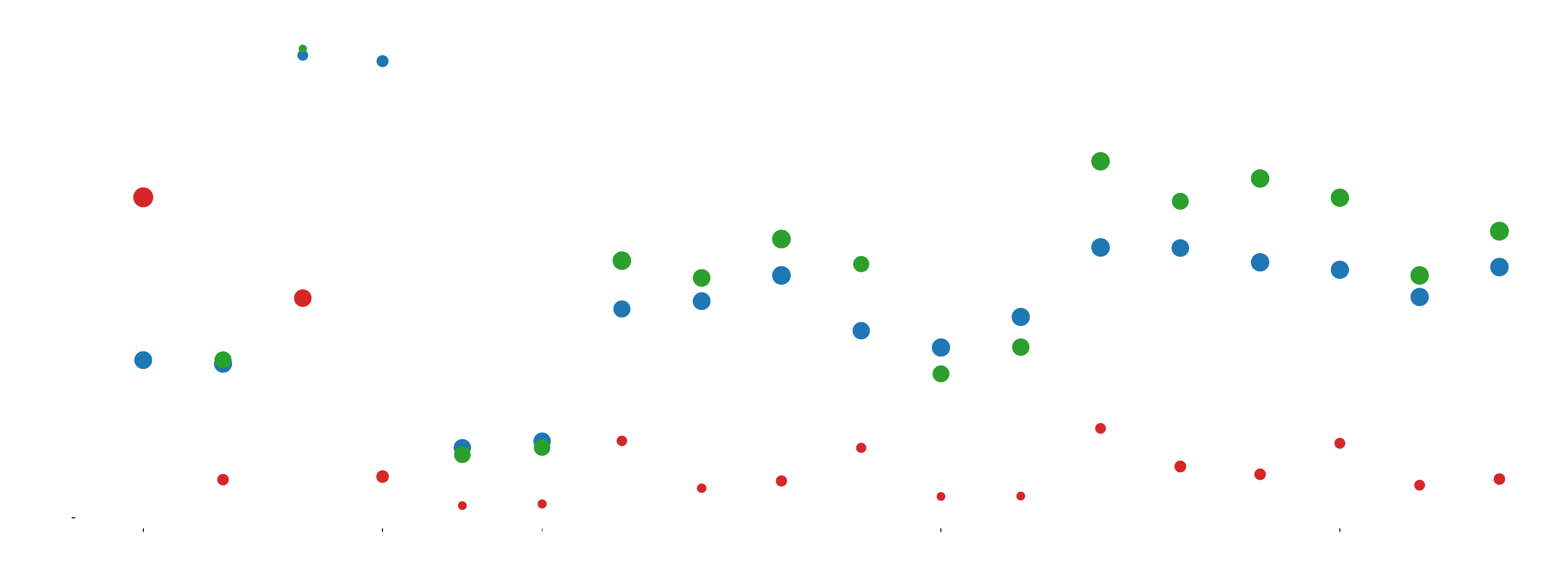
	\caption{Map quality evaluation on the \helipr{} dataset averaged over the three recording instances in each scenario across all three \lidar{}s. We show a scatter plot of the average inlier RMS scores for the nearest-neighbor distances between local maps at revisit locations. The size of the markers represent the magnitude of the standard deviation. The inlier threshold is set to $1.0$\,m.}
	\label{fig:map_quality_helipr}
\end{figure*}

\begin{figure}[t]
	\centering
	\def\svgwidth{0.99\linewidth}
	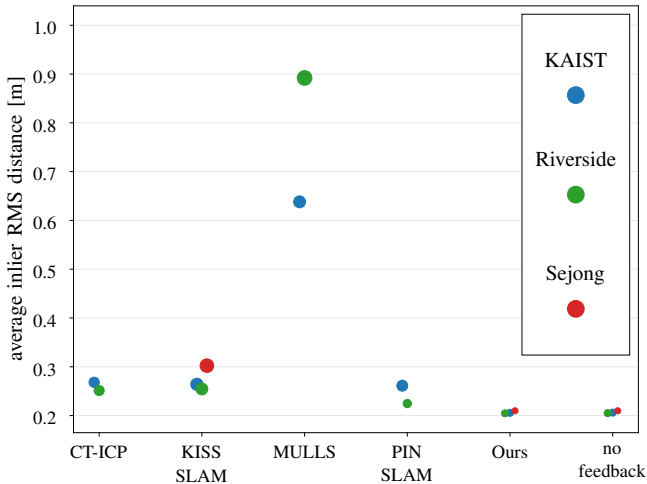
	\caption{Map quality evaluation on the \mulran{} dataset averaged over the three recording instances in each scenario. We show a scatter plot of the average inlier RMS scores for the nearest-neighbor distances between local maps at revisit locations. The size of the markers represent the magnitude of the standard deviation. The inlier threshold is set to $1.0$\,m.}
	\label{fig:map_quality_mulran}
  \vspace{-10pt}
\end{figure}

Trajectory-based metrics alone do not fully capture the quality of the reconstructed map. We therefore complement the conventional evaluation with a quantitative assessment of local map consistency at revisit locations. Our protocol measures the geometric consistency between independently reconstructed maps of the same physical region acquired during different traversals. The results suggest that our approach enhances the local map accuracy at revisit locations.

We first identify revisit locations from the reference trajectory by projecting it onto a coarse grid in the xy-plane. Within each occupied cell, we extract contiguous trajectory segments. We retain cells containing multiple such segments as revisit regions, and store the associated trajectory segments with the corresponding scan indices.

To evaluate each SLAM baseline, we use the stored scan indices to construct a local map for every trajectory segment within a revisit cell based on the estimated global poses, yielding multiple local maps representing independent observations of the same region. We compare the local maps directly without any additional registration. We compute the root mean square~(RMS) distances between nearest-neighbor point correspondences between every pair of local maps within each revisit cell. We aggregate the RMS values across all revisit cells and report their mean and standard deviation as a measure of map consistency.

We first discuss the map quality evaluation results for the \helipr{} dataset. \figref{fig:map_quality_helipr} shows that our method occupies the lower regions of the plots, with lower mean RMS error and therefore better geometric agreement between independent revisits. This trend is consistent across all environments and all three different \lidar{} sensors. The ablation study in the last column shows the results when we disable the retroactive loop closure feedback. We see a slight increase in RMS distances across all sequences and sensors, confirming its contribution to improving local map consistency.

The Bridge sequences remain the most challenging scenarios because they contain long traversals through perceptually aliased environments. Nevertheless, our method achieves substantially lower RMS errors than all baselines. Combined with the trajectory evaluation, these results indicate that our approach remains robust in perceptually aliased environments, and simultaneously provides more consistent local reconstructions. The evaluation also reveals a clear dependence on \lidar{} sensor characteristics. Across all methods, the spinning Ouster sensor~(red) consistently achieves lower RMS errors and lower variance than the Aeva~(green) and Avia~(blue) sensors, suggesting that a wider field-of-view and denser scan pattern provides better reconstruction consistency.

We observe a similar performance trend on the \mulran{} dataset~(\figref{fig:map_quality_mulran}), where our method again achieves the lowest mean RMS error across all recording sessions. Since these recordings are not affected significantly by perceptual aliasing, our front-end loop-closure module does not miss many loops for the latter retroactive loop-closure stage to recover. Therefore, the ablation study~(last column) does not show a perceivable difference in map quality. This observation is consistent with the \ac{ate} scores on these sequences as reported in~\tabref{tab:ApolloMulRan}.

Furthermore, we are the only baseline alongside KISS-SLAM to succeed on the Sejong sequences which are recorded on a highway with a little overlap in the trajectory. We obtain a better map quality in comparison to KISS-SLAM to complement the best \ac{ate} score. This demonstrates that the improvements generalize across different datasets without any parameter tuning.

Interestingly, several methods exhibit comparable \ac{ate} values but have noticeably different map consistency scores. This highlights that trajectory accuracy alone is insufficient to characterize SLAM performance and motivates the proposed revisit-based evaluation.

In summary, our proposed \ac{slam} approach achieves competitive global trajectory accuracy while improving local map consistency at revisit locations. The results demonstrate that information-aware pose graph optimization along with hierarchical and retroactive loop closure improves both trajectory estimation and map quality.

\section{Conclusion}\label{sec:conclusion}
In this work, we presented a novel and effective 3D \lidar{} \ac{slam} approach that improves global trajectory estimation and local map quality through geometry-aware odometry information estimation, hierarchical multi-resolution loop closure, and geometry-based retroactive loop closure detection. Experimental results obtained on several challenging benchmark datasets demonstrate competitive trajectory accuracy together with consistently better map quality at revisit locations compared with state-of-the-art \ac{slam} systems.

\bibliographystyle{plain_abbrv}
\bibliography{glorified,new}
\end{document}